\documentclass[lettersize,journal]{IEEEtran}
\usepackage{amsmath,amsfonts}
\usepackage{array}
\usepackage[caption=false,font=normalsize,labelfont=sf,textfont=sf]{subfig}
\usepackage{textcomp}
\usepackage{stfloats}
\usepackage{url}
\usepackage{verbatim}
\usepackage{graphicx}
\usepackage{amssymb}
\usepackage[most]{tcolorbox}
\usepackage[switch]{lineno} 

\usepackage{cite}
\usepackage{booktabs}
\usepackage{xspace}
\usepackage{hyperref}

\usepackage{multicol}
\usepackage{multirow}
\usepackage{algorithm}
\usepackage{algpseudocode}
\PassOptionsToPackage{implicit=false}{hyperref}
\hypersetup{hidelinks,
	colorlinks=true,
	allcolors=black,
	pdfstartview=Fit,
	breaklinks=true}
\definecolor{lime}{HTML}{A6CE39}
\DeclareRobustCommand{\orcidicon}{
\begin{tikzpicture}
\draw[lime, fill=lime] (0,0)
circle[radius=0.16]
node[white]{{\fontfamily{qag}\selectfont \tiny \.{I}D}};
\end{tikzpicture}
\hspace{-2mm}
}
\foreach \x in {A, ..., Z}{%
\expandafter\xdef\csname orcid\x\endcsname{\noexpand\href{https://orcid.org/\csname orcidauthor\x\endcsname}{\noexpand\orcidicon}}
}

\usepackage{xspace}
\newcommand{\ie}{{\emph{i.e.}}\xspace}
\newcommand{\eg}{{\emph{e.g.}}\xspace}

\usepackage{soul}
\soulregister{\citep}7  
\soulregister\cite7
\soulregister{\ref}7 

\usepackage{xcolor}
\usepackage{soul}
\newcommand{\mathcolorbox}[2]{\colorbox{#1}{$\displaystyle #2$}}

\begin{document}
\title{Embodied Perception for Test-time Grasping Detection Adaptation with Knowledge Infusion}

\author{Jin Liu\hspace{-1.5mm}\orcidB{}, Jialong Xie\hspace{-1.5mm}\orcidC{}, Leibing Xiao\hspace{-1.5mm}\orcidE{}, Chaoqun Wang\hspace{-1.5mm}\orcidA{}, Fengyu Zhou\hspace{-1.5mm}\orcidD{}
\thanks{ Jin Liu, Jialong Xie, Fengyu Zhou, and Chaoqun Wang are with the School of Control Science and Engineering, Shandong University, China. Leibing Xiao is with the Unmanned Systems Technology Research Institute, Northwestern Polytechnical University, China.}
\thanks{Email: \{202120638,202220703\}@mail.sdu.edu.cn,x13033968676@mail\ .nwpu.edu.cn,\{chaoqunwang,zhoufengyu\}@sdu.edu.cn.}}


\maketitle

\begin{abstract}
It has always been expected that a robot can be easily deployed to unknown scenarios, accomplishing robotic grasping tasks without human intervention. Nevertheless, existing grasp detection approaches are typically off-body techniques and are realized by training various deep neural networks with extensive annotated data support. {In this paper, we propose an embodied test-time adaptation framework for grasp detection that exploits the robot's exploratory capabilities.} The framework aims to improve the generalization performance of grasping skills for robots in an unforeseen environment. Specifically, we introduce embodied assessment criteria based on the robot's manipulation capability to evaluate the quality of the grasp detection and maintain suitable samples. This process empowers the robots to actively explore the environment and continuously learn grasping skills, eliminating human intervention. Besides, to improve the efficiency of robot exploration, we construct a flexible knowledge base to provide context of initial optimal viewpoints. Conditioned on the maintained samples, the grasp detection networks can be adapted in the test-time scene. When the robot confronts new objects, it will undergo the same adaptation procedure mentioned above to realize continuous learning. Extensive experiments conducted on a real-world robot demonstrate the effectiveness and generalization of our proposed framework. 
\end{abstract}

\begin{IEEEkeywords}
Test-time adaptation, embodied perception, grasp detection, knowledge retrieval 
\end{IEEEkeywords}

\section{Introduction}
Service robots are increasingly engaged in daily service tasks, where grasp detection has emerged as a critical step for task completion. To endow robots with more accurate grasp detection ability, current studies mainly focus on designing deep learning networks~\cite{jiang2011efficient}\cite{yu2022se}, where they are trained with vast amounts of annotated data from human experts. These deep learning methods suffer from severe performance degradation when faced with unforeseen scenarios. Moreover, the data collection process is both time-consuming and labor-intensive for non-expert users, leading to impediments in the widespread deployment of robotics. To this end, appropriately adapting one pre-trained grasp detection network with the unlabelled data from vision sensors is a meaningful way to guarantee the robot's performance in daily household tasks. 

In recent developments within the image classification domain, test-time adaptation techniques~\cite{wang2021tent,chen2022contrastive,chen2023improved} have been proposed to tackle the above adaptation challenges in grasp detection without extra laboring annotations. Specifically, these test-time adaptation techniques aim to adapt pre-trained networks to new environments using online unlabeled test samples. However, directly transferring these test-time adaptation techniques into the grasp detection task can not be immediately finalized. Firstly, they are originally designed for classification tasks, where the predictions are often object-level. In contrast, the grasp detection methods~\cite{yu2022se}\cite{liu2023continual} that we utilize in test-time adaptation entail pixel-level predictions. Besides, their samples are limited to a single perspective, failing to leverage the embodied ability of the robot to acquire more appropriate samples.

\begin{figure}
    \centering
    \includegraphics[width=\linewidth]{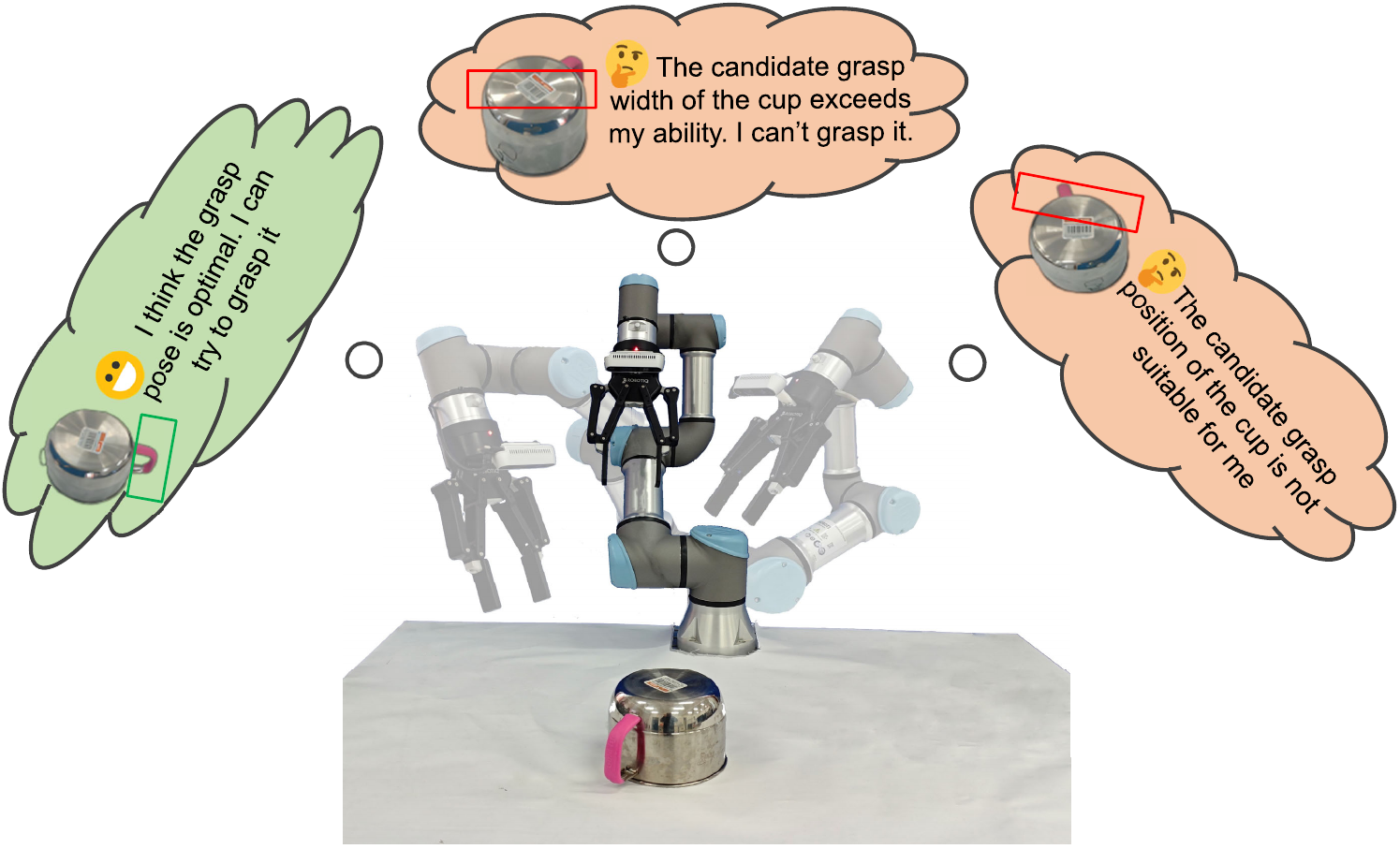}
    \caption{An example of embodied test-time grasp detection for robotics, where the robot can only access the unlabelled data from unseen scenes with one pre-trained grasp detection network. The green rectangle indicates a viable grasping posture, whereas the red rectangle indicates an unsuccessful one.}
    \label{fig:figure1}
\end{figure}

To assimilate the test-time adaption technique in grasp detection, this paper investigates the robot exploration capability and presents an embodied perception pipeline. The embodied perception empowers the robot to actively explore and maintain suitable samples while accounting for its physical limitations and abilities, thereby offering a practicable solution. As illustrated in Fig.~\ref{fig:figure1}, during the process of grasping execution, the robot can assess candidate grasping results from multiple viewpoints. Thus, it can guarantee the successful completion of the task. To this end, we pre-distribute a set of fine-grained view candidates~\cite{morrison2019multi}\cite{breyer2022closed} and organize them into coarse-grained observation groups. Additionally, we present an embodied assessment criterion regarding the robot's physical parameters to evaluate the anticipated quality of candidate grasping postures. This criterion aims to retain proficient samples with pseudo labels to address the challenges in test-time adaptation for grasp detection. Notably, to further improve the efficiency and quality of the exploration, we enable the robot to access a knowledge pool of optimal viewpoints for the potential object-grasping posture through a knowledge retrieval network. The preserved samples with pseudo labels are utilized to adapt both the grasp detection network and the knowledge retrieval network to the current scene. These optimized networks are deployed and optimized again with encountered objects to enable the robot to work continuously.

The main contributions of this paper are summarized as follows:
\begin{itemize}
    \item We propose an embodied test-time adaptation framework for robotic grasping detection. The framework empowers the robot to continuously adapt the pre-trained grasp detection network to an unfamiliar environment without human intervention. 
    \item We propose an embodied assessment criterion regarding the robot's physical parameters to better determine the credibility of the current grasping strategy and preserve the qualified samples for test-time adaptation.
    \item We design a knowledge retrieval module for the robot to access the knowledge of optimal viewpoints from the knowledge pool. Besides, this module enables the robot to conduct efficient, high-quality exploration along pre-distributed viewpoints.
\end{itemize}

\section{Related Work}
Our proposed embodied test-time adaptation framework for grasp detection seeks to enable the robot to autonomously acquire grasping postures based on its inherent perception and decision-making abilities, eliminating the need for human intervention in dynamically changing environments. The relevant studies to our work are categorized into the following three groups:

1) \textit{Grasp Detection with Deep learning}, {which aims to predict optimal grasping poses for the objects using deep learning techniques, is a prominent focus within the field of robotic manipulation~\cite{cheng2023anchor,TwinDec,liu2023continual}. The work~\cite{cheng2023anchor} is one of the classic works that designs an anchor-based two-stage grasp detection network for the robot with a parallel gripper to grasp novel objects. Nevertheless, such a two-stage approach cannot balance efficiency and accuracy well. Toward this problem, many recent works utilize convolution and deconvolution networks to formulate a pixel-wise grasp detection network to achieve higher efficiency and accuracy~\cite{TwinDec,liu2023continual}. Currently, these methods heavily rely on data-driven approaches involving laboring human annotations, encountering challenges in acquiring proficiency in grasp detection when confronted with unlabeled data in dynamic environments.}
2) \textit{Test-time adaptation}, aiming to adapt a pre-trained model to a potentially altered target test domain, has been extensively employed in image classification~\cite{wang2021tent}. {In this scenario, the model will be optimized relying solely on the acquired samples without any human annotations~\cite{chen2023improved}. Nevertheless, there have been limited studies applied to the realm of robotics. As an earlier attempt, Zhang et al.~\cite{zhang2023unseen} formulate a non-parametric entropy objective to conduct the test-time adaptation to enable the robot to segment unseen objects in the real-world scenario. Similarly, Li et al.~\cite{Monocular} propose to combine both self-training of the supervised branch and pseudo labels from the self-supervised branch to attain a test-time domain adaptation for monocular depth estimation. However, these methods solely focus on the characteristics of image alterations in new environments, overlooking the holistic integration of the robot's embodied exploration capabilities for domain adaptation.}

3) \textit{Active perception}, enabling the robots to actively explore in ever-changing scenes~\cite{chaudhary2023active,wei2023discriminative,chaplot2021seal}, has garnered increasing attention. {Chaudhary et al.~\cite{chaudhary2023active} develop an active perception system aimed at enhancing the robotic capability for object segmentation in simulated cluttered scenes, which is achieved through adjustments in the robot's 3D position and the implementation of deep reinforcement learning. Nevertheless, the quality of the collected data is difficult to guarantee. Subsequently, Wei et al.~\cite{wei2023discriminative} propose a discriminative active learning framework to utilize labeled and unlabelled data for robotic grasping detection by feature clustering and human estimation. However, the cost of human intervention is high. In contrast to their settings, we propose leveraging the robot's embodied perception capabilities to explore and maintain optimal grasping postures actively. Consequently, it can adapt to new test-time scenes without any human intervention.}

\section{Problem statement}
In real-world robotic applications, robots inevitably perform grasping tasks in unforeseen environments. Nevertheless, the large differences across domains result in the degradation of the generalization ability. To alleviate this issue, in this paper, we seek to study how to improve the generalization performance of grasping skills for robots in new environments based on their own capabilities without any newly annotated samples, which is also termed embodied test-time adaptation for grasp detection. 

{Without the loss of generality}, let GNet$_{\theta}$ be a grasp detection network that is pre-trained on labeled grasping samples $D_s=\{(x_i,y_i)\}_{i=1}^{N_s}$ with optimized parameter $\theta$. Embodied test-time adaptation (ETA) aims to boost the network performance with samples from active exploration in new scenes conditioned on the robot's embodied ability. We regard the learning objective as typical smooth L1 loss, which is defined as:
\begin{equation}
    \label{oploss}
    \text{min}_{\theta} \frac{1}{N_a}\sum_{a=1}^{N_a}\text{L1}((x_a,y_a);\theta),
\end{equation}
where $N_a$ denotes the number of samples $x_a$ with pseudo-labels $y_a$ collected via active exploration.

\section{Methodology}
\subsection{Overview}
\begin{figure*}
    \centering
    \includegraphics[width=\linewidth]{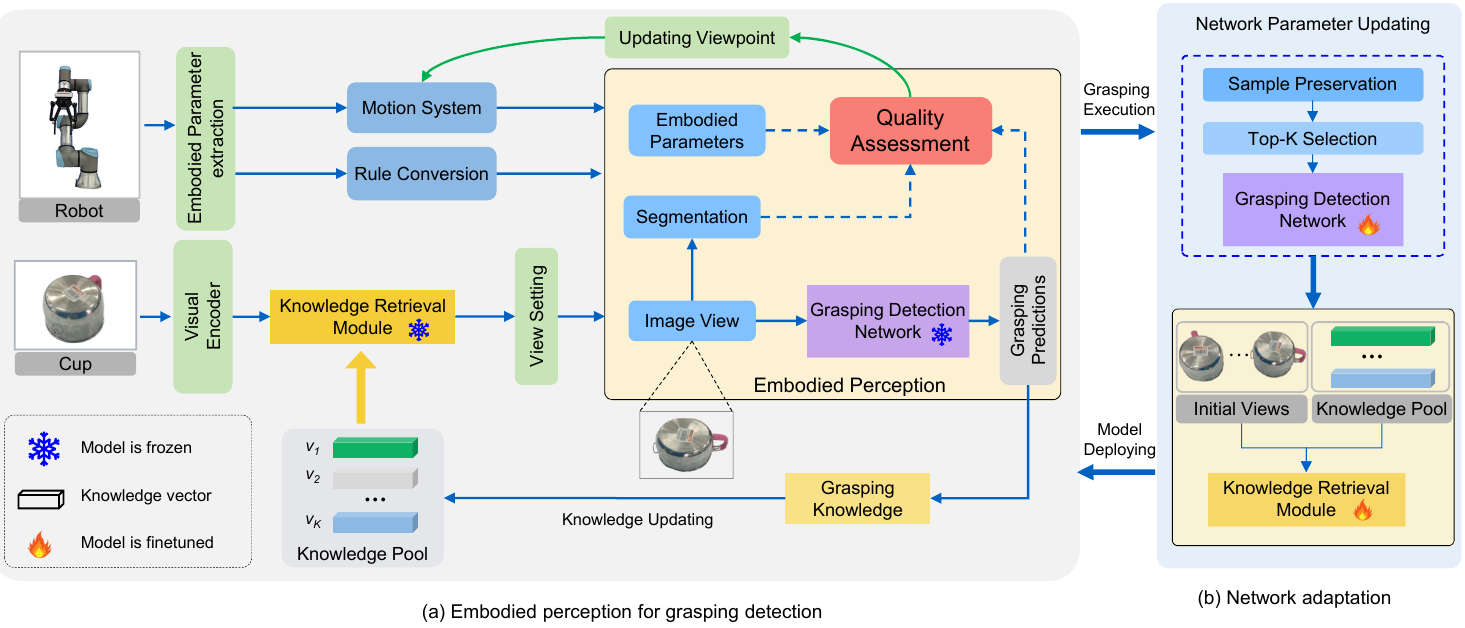}
    \caption{An overview of the proposed embodied test-time adaptation framework for grasp detection. The robot first retrieves the historical grasping knowledge related to the optimal candidate viewpoint. Then, it actively explores different viewpoints and preserves optimal samples based on embodied assessment indicators. Finally, conditioned on the collected samples, the knowledge retrieval network and grasp detection network are optimized. These optimized networks during test time are deployed in the current scene to facilitate scene adaptation. }
    \label{fig:model}
\end{figure*}

The overview of the proposed framework is illustrated in Fig.~\ref{fig:model}. It mainly consists of three components:{ (1) Grasping knowledge retrieval module for retrieving historical knowledge related to the optimal candidate viewpoint for the grasp manipulation; (2) Embodied perception module for actively exploring different viewpoints to preserve appropriate grasp postures based on the embodied assessment indicator; (3) Network optimization module for updating the network parameters including knowledge retrieval network and grasp detection network based on the knowledge pool and preserved high-quality samples, respectively.} The details are described in the following sections.
\subsection{Grasping Knowledge Retrieval Module}
When humans attempt to grasp an object, they initiate a knowledge retrieval sequence to determine the optimal initial viewpoint for efficient and effective object manipulation, rather than from the inefficient method of commencing with the initial observation point and contemplating the grasping strategy. Inspired by such a process, we propose the knowledge retrieval model to acquire grasping knowledge related to the viewpoints. Firstly, we utilize one pre-trained visual encoder ResNet~\cite{he2016deep} to extract the semantic context features $f_{\text{I}}$ from the captured RGB image $\text{I}_{RGB}$:
\begin{equation}
    \label{resnet}
     f_{\text{I}}=\text{ResNet}(\text{I}_{RGB})\in \mathbb{R}^{1\times D},
\end{equation}
{where $D$ denotes the dimension of the feature representation.} Subsequently, We retrieve relevant knowledge of the extracted features $f_{\text{I}}$ from the knowledge pool, which is composed of compressed knowledge vector embeddings $V=\{v_1,\cdots,v_K\}$. The retrieval process can be computed by the following cos similarity:
\begin{equation}
    \label{retrieval}
    \delta=\text{SIM}(f_I,V)=\frac{f_I V^T}{||f_I||\cdot ||V||}\in \mathbb{R}^{(0,1)},
\end{equation}
where $\delta$ denotes $K$ similarity values. {SIM(A,B) represents the similarity calculation function between A and B.} If all the similarity values are less than 0.95, the object will be regarded as brand new and unseen. Conditioned on $\delta$, we can obtain the top relevant knowledge embedding $F_\text{know}$. Then, we propose an observation prediction network OPNet to acquire the potential initial observation position in the viewpoint trajectory. {The OPNet mainly consists of two fully connected layers (MLPs) and takes the knowledge embeddings as inputs}, which can be defined as:
\begin{equation}
    \label{viewpredict}
       P=\text{MLPs}(\text{ReLU}(F_\text{know})))\in \mathbb{R}^{1\times O},
\end{equation}
{where MLPs denote the fully connected layers. ReLU represents the activation function. $O$ denotes the number of viewpoints. $P$ denotes the logits of the viewpoint prediction.} Finally, we select the key observation $k$ with the highest probability value as the initial observation position. The process can be computed:
\begin{equation}
    \label{sort}
    k=\text{argmax}(\text{Softmax}(P)).
\end{equation}

\subsection{Embodied Perception Module}
Embodied perception empowers the robot to actively explore the viewpoints along a predefined trajectory and maintain suitable samples. The process is outlined in Algorithm~\ref{alg:alg1}, and the details are described below.

\begin{figure}[t]
    \centering
    \includegraphics[width=.8\linewidth]{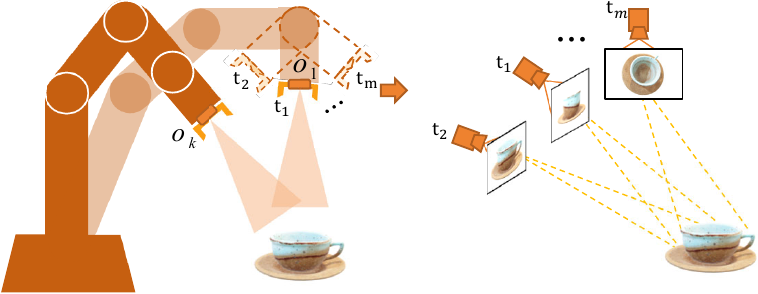}
    \caption{Examples of active explorations. The figure on the left illustrates the observation positions, whereas the one on the right shows the viewpoints.}
    \label{fig:observes}
\end{figure}
\begin{algorithm}[t]
\caption{Active Exploration}\label{alg:alg1}
\begin{algorithmic}[1]
\Require Embodied parameters $R_e$, viewpoint trajectory $\mathcal{T}$ and $\mathcal{O}$, grasp detection network GNet, convex hull conversion function CH, quality assessment function QA.

\State $o_{f} \Leftarrow \mathcal{O}$

\State $OV=\{o_f,t_{3f-1},\cdots,t_{\text{V}}\},t_{*}\in \mathcal{T}$\Comment{New viewpoints}
\For {$v\in OV$}
    \State $(I_{RGB},I_{D}) \Leftarrow \mathcal{O}_f$ \Comment{Retrieving images}
    \State $\mathcal{G}^v\Leftarrow$ GNet($(I_{RGB}^{v},I_{D}^v)$)
    \If{not $\mathcal{G}^v$ satisfies $R_e$}
        \State Updating viewpoint
        \State Execute line 3
    \EndIf
    \State $\mathcal{M}^v \Leftarrow  \text{CH}(\text{SAM}(\text{I}_{rgb}^v))$
    \State $\mathcal{S}^v\Leftarrow \text{QA}( R_e,~\mathcal{M}^v,\mathcal{G}^v)$
    \If {$\mathcal{S}^v\ge \epsilon$}
        \State Execute grasping action
        \State Break
    \Else
        \State Updating viewpoint
        \State Execute line 3
    \EndIf   
\EndFor
\State Model parameter updating
\end{algorithmic}
\end{algorithm}

\paragraph{Pre-distributed Viewpoints}
\label{viewpoints01}
Similar to previous works~\cite{breyer2022closed}\cite{morrison2019multi}, we pre-define a viewpoint trajectory which consists of V fine-grained discrete viewpoints $\mathcal{T}=\{t_1, t_2, \cdots, t_v \}_{v=1}^{\text{V}}$ and organize them into four coarse-grained observation position groups $\mathcal{O}=\{o_1,\cdots,o_k\}_{k=1}^{K}$, where each observation consists of $m$ viewpoints, as is illustrated in Fig.~\ref{fig:observes}. We propose to embrace an active exploration strategy that progresses from a broad overview to a detailed examination. Specifically, the robot first moves to the observation position $o\in \mathcal{O}$ based on the retrieved knowledge and then conducts detailed exploration following the pre-distributed viewpoints $t\in \mathcal{T}$. 

\paragraph{Grasp Detection}
In this part, we focus on how to predict the candidate grasping rectangle conditioned on the RGB and depth image inputs by a pre-trained grasp detection network GNet:
\begin{equation}
    \label{grasppredict}
     \mathcal{G}=\text{GNet}_{\theta}(\text{I}_{RGB},\text{I}_{D}),\quad \theta_{t+1}\Leftarrow \theta_{t},
\end{equation}
where $\mathcal{G}$ denotes grasping pose. Notably, in this process, the grasp detection network is frozen and only utilized to predict the grasping rectangles. Following the general definition of the grasping pose for one input 2D image in previous works~\cite{zhou2023aagdn,yu2022se}, we represent the grasping rectangle as $ g=\{x, y, w, \phi, q\}\in \mathcal{G}$, 
where $(x,y)$ denotes the coordinates of the center point of the grasping rectangle, $w$ denotes the opening width of the parallel-plate gripper~\footnote{https://robotiq.com/products/2f85-140-adaptive-robot-gripper} with the range of $[0,\text{w}_{max}]$, $\phi$ is the rotation angle of the grasping rectangle around Z-axis with the range of $[-\frac{\pi}{2},\frac{\pi}{2}]$, $q$ is the quality score of the grasping rectangle. Conditioned on the quality score, we can obtain the best grasping rectangle candidate by $g^*=\text{argmax}_{q}\mathcal{G}$    

\begin{figure}
    \centering
    \includegraphics[width=.9\linewidth]{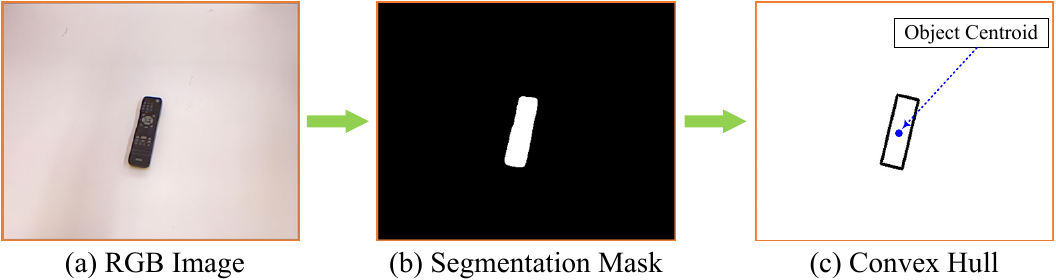}
    \caption{An example of the process of obtaining the convex hull and object centroid.}
    \label{fig:maskhull}
\end{figure}

\paragraph{Image Segmentation}
Since the predicted grasping postures inevitably possess imprecise results, we introduce semantic segmentation~\cite{xie2023satr} to obtain the grounding mask for each referenced object to assist the fine-grained quality assessment process. In this paper, we utilize the state-of-the-art segmentation model, \ie Segment Anything (SAM)~\cite{kirillov2023segment}, for the offline object mask generation without human annotations. The process can be formulated as:
\begin{equation}
    \label{seg}
     \text{I}_{mask}^v=\text{SAM}(\text{I}_{rgb}^v),\quad v\in \mathcal{T},
\end{equation}
where $\text{I}_{rgb}^v$ denotes the image from viewpoint $v$. Conditioned on grounding mask, we then construct a convex hull $\mathcal{M}^v$ by Opencv tools~\footnote{https://opencv.org/} for each object~\cite{vuong2023grasp} as a step towards subsequent quality assessment, which is illustrated in Fig.~\ref{fig:maskhull}. 
\paragraph{Embodied Parameters}
When a robot is preparing to engage in object-grasping manipulation, it needs to take into its own embodied parameters. In this paper, we consider the following three important principles derived from the previous works~\cite{cheng2023anchor}\cite{yu2022skgnet} for the subsequent quality assessment procedure. 


\begin{tcolorbox}[colback=yellow!35,
                  colframe=black,
                  width=\linewidth,
                  arc=1mm, auto outer arc,
                  boxrule=0.5pt,
                 ]
$\blacktriangleright$ $ w_r \in [\frac{\text{W}_{max}}{10},\frac{\text{W}_{max}}{4}], w_r\in \mathcal{G}_r$, where $\text{W}_{max}$ denotes the maximum opening width of the real parallel-plate gripper. 

$\blacktriangleright$ $z_r\in [\text{CL}_{min},\text{CL}_{max}]$, where $z_r$ is the detected distance between the desired object and the end of the parallel-plate gripper. $\text{CL}_{*}$ denotes the operational range for camera depth detection.

$\blacktriangleright$ $ p_r=(x,y,z) \in R_m, p_r\in \mathcal{G}_r$, where $R_m$ denotes the coordinate points within three-dimensional space accessible to robot space.
\end{tcolorbox}


\paragraph{Quality Assessment}

After obtaining the predicted grasping postures, the robot should have the ability to self-evaluate and achieve outcome evaluation without human intervention. Conditioned on the embodied parameters, semantic segmentation, and grasping detection results, we propose the following criteria to assess whether the current predicted grasping postures can facilitate the robot to successfully grasp the object. Notably, the results that do not satisfy the embodied parameter constraints have been excluded. 
\begin{itemize}
    \item For all center points of the predicted grasping rectangle candidates at the current viewpoint, it should fall on the object to make a stable grasping. Formally, the process can be defined as :
    \begin{equation}
        \label{grasp01}
        \begin{split}
             \mathcolorbox{yellow}{q^v=(x_c^v,y_c^v), \quad \forall v\in\mathcal{T},\ q^v\in \mathcal{G}^v,}\\
        \mathcolorbox{yellow}{s.t. \quad (x_c^v,y_c^v)\cap \text{I}^v_{mask} \neq \emptyset,\qquad}
        \end{split}
    \end{equation}
    \hl{where $v$ denotes the viewpoint from the set $\mathcal{T}$. $q^v=(x_c^v,y_c^v)$ denotes the the center of grasping detection pose at viewpoint $v$. $\mathcal{G}^v$ represents the set of grasping poses. Besides, $q^v$ should satisfy the condition, $q^v\cap \text{I}^v_{mask} \neq \emptyset$, to achieve stable grasping, where I$_{mask}^v$ represents the segmented regions of the object at viewpoint $v$.}
    \item For all the center points, it should be close to the object centroid that is calculated from the convex hull. Formally, the process can be formulated as :
    \begin{equation}
        \label{grasp02}
        \underset{i\in 1,2\cdots, \text{N}}{\text{Minimize}}(||q^v_i-m^v||_2),\quad m^v=(x_{mc},y_{mc})\in \mathcal{M}^v,
    \end{equation}
    where N denotes the number of predicted grasping candidates.
    \item The predicted opening width of the gripper should be as small as possible. Besides, the end point of the gripper should not fall on the object. Formally, the process can be defined as :
    \begin{equation}
        \label{grasp03}
            \underset{i\in 1,2\cdots, \text{N}}{\text{Minimize}}(||w_i^v||_2), \quad w_i \in \mathcal{G}^v,
    \end{equation}
    where the coordinate of the four vertices $x^e$ for the predicted grasping rectangle must satisfy:
    \begin{equation}
        \label{grasp031}
         \forall (x^e_r,y^e_r)\cap \mathcal{M}^v = \emptyset,\quad r\in 1,\cdots,4.
    \end{equation}
    \end{itemize}
Notably, Eq.~\ref{grasp01} and Eq.~\ref{grasp03} are the primary criteria that all the predictions should satisfy for evaluating the results. Therefore, once the anticipated grasping outcomes align with the aforementioned requisites, the quantifiable criteria for evaluation are specified as follows:
\begin{equation}
    \label{finalgrasp}
    \mathcal{S}=\frac{\lambda_1}{||q^v_i-m^v_i||_2}+ \frac{\lambda_2}{||w_i^v||_2},\quad i\in 1,\cdots,N,v\in \mathcal{T},
\end{equation}
where $\lambda_1$ and $\lambda_2$ are trade-off parameters. In this paper, we set $\lambda_1$ to 90 and $\lambda_2$ to 122. Subsequently, we manually set one grasping execution threshold $\delta$ with the following criteria:
\begin{equation}
    \label{shouldgrasp}
    Action=
    \begin{cases}
          \texttt{Grasp}(object),  & \text{if}\quad \mathcal{S}\ge \epsilon \\
         \texttt{Explore}(viewpoint), & \text{otherwise}
    \end{cases}
\end{equation}
Notably, we empirically set the threshold $\epsilon$ to 4.0 for all the experiments.
\subsection{Network Optimization Module}
Conditioned on the samples acquired through embodied perception, we paired them with candidate prediction rectangles that satisfy the quality criteria:
\begin{equation}
    \label{select}
       \{s_{a1},s_{a2},\cdots,s_{aM}\}=\text{Pair}(\text{I}_{RGB},\text{I}_{D},\mathcal{G}^a),
\end{equation}
where the sample $s_{a*}$ consists of one RGB image $\text{I}_{RGB}$, one depth image $\text{I}_{D}$, and active labeled annotations $\mathcal{G}^a$. Subsequently, we aim to adapt the pre-trained grasp detection network with these active samples to the current new scene. In this process, the inputs $\{(\text{I}_{RGB}^1,\text{I}_{D}^1),(\text{I}_{RGB}^2,\text{I}_{D}^2),\cdots,(\text{I}_{RGB}^M,\text{I}_{D}^M)\}$ are utilized to generate the grasping rectangles $ y_p=\mathcal{G}_p=\text{GNet}(x^a)$ and $y^a=\{ g_1^a,g_2^a,\cdots,g_M^a \}$ are set as the labels. The goal is to minimize the differences between $y_p$ and $y_a$ by the following Smooth L1 loss~\cite{zhou2023aagdn}:
\begin{equation}
    \label{smoothl1}
    \mathcal{L}_{act}=\frac{1}{M}\sum_{m=1}^{M} 
    \begin{cases}
          0.5(y^a_m-y^p_m)^2,  & \text{if $|y^a_m-y^p_m|<1$} \\
         |y^a_m-y^p_m|-0.5, & \text{otherwise}
    \end{cases}
\end{equation}
After adaptation, the grasp detection network will be frozen again for the subsequent active grasping process. Simultaneously, conditioned on the quality assessment scores, we select the optimal grasping observations $o_{best}\in \mathcal{O}$ to grasp the object and preserve the image from the initial viewpoint $t_1$. Then, we utilize the same feature extraction process in Eq.~\ref{resnet} to formulate new knowledge vectors $F_{\text{know}}$ and update the knowledge pool. With the newly updated knowledge pool and initial views $\text{I}_{t_1}$, we update the OPNet in the knowledge retrieval model by the following cross-entropy loss function:
\begin{equation}
    \label{knowupdate}
    \begin{split}
        F_{\text{know}} \Longleftarrow \text{Update}(F_{\text{know}},\text{ResNet}(\text{I}_{t_1})),\\
      \mathcal{L}_{know}=\text{CE}(\text{OPNet}(F_{\text{know}}),\text{argmax}(\mathcal{O})),     
    \end{split}
\end{equation}
Therefore, the overall optimization loss during test-time adaptation is formulated as:
\begin{equation}
    \label{overall}
    \mathcal{L}=\mathcal{L}_{act}+\mathcal{L}_{know}.
\end{equation}



\section{Experiments and Results}
\label{exp}

\subsection{Experimental Settings}
\begin{figure}
    \centering
    \includegraphics[width=0.8\linewidth]{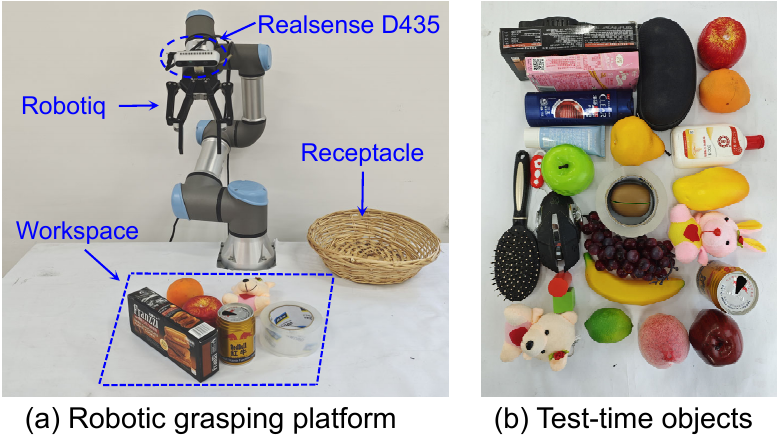}
    \caption{(a) Overview of the robotic grasping platform. (b) Objects utilized in test-time adaptation.}
    \label{fig:platform}
\end{figure}

\paragraph{Implementation Details}
We conduct all the experiments on Ubuntu 18.04 with a single NVIDIA RTX 3090. The baseline grasp detection network GGCNN~\cite{morrison2018closing} is first pre-trained on the Cornell dataset~\cite{jiang2011efficient}. Then, the pre-trained grasp detection network is optimized during test time following the steps in Fig.~\ref{fig:model}. The experimental robotic platform and 27 kinds of test-time objects are shown in Fig.~\ref{fig:platform}. After being fully optimized, the robot can continuously learn the grasping operation of unknown objects in new scenarios.
\paragraph{Evaluation Metrics}
We adopt the widely utilized metric to evaluate the quality of the antipodal grasp. Specifically, the predicted grasping postures that are satisfied the following two conditions can be considered optimal, \ie The angle difference between the prediction and the ground truth is less than 30$^{\circ}$ and the Jaccard index that indicates the intersection over union (IoU) between the predicted grasping rectangle $P_g$ and the ground-truth $G_g$ is larger than 0.25, which is calculated as:
\begin{equation}
    J(P_g,G_g)=\frac{P_g\cap G_g}{P_g\cup G_g}.
\end{equation}
Notably, during the initial phase of the active perception process, the robot will save the RGB-D images of each object. Subsequently, three experts are enlisted to annotate grasping rectangles for each object, and additional three discriminators are tasked with voting for the most optimal annotations. This enables us to employ the above criteria in evaluating both our proposed methods and the comparative methods.

\hl{Besides, we propose execution time efficiency (EE) to evaluate the efficiency of active exploration. Formally, it is defined as follows:}
\begin{equation}
    \label{explore}
     \mathcolorbox{yellow}{\text{EE}=\frac{1}{SG}\cdot 100\%,\quad SG\in 1,2,\cdots, \text{V},}
\end{equation}
\hl{where $SG$ denotes the number of exploration steps required to successfully grasp an object. Notably, in this paper, we consider each step to incur an identical cost.}
\subsection{Main Results}
\begin{figure}
    \centering
    \includegraphics[width=.85\linewidth]{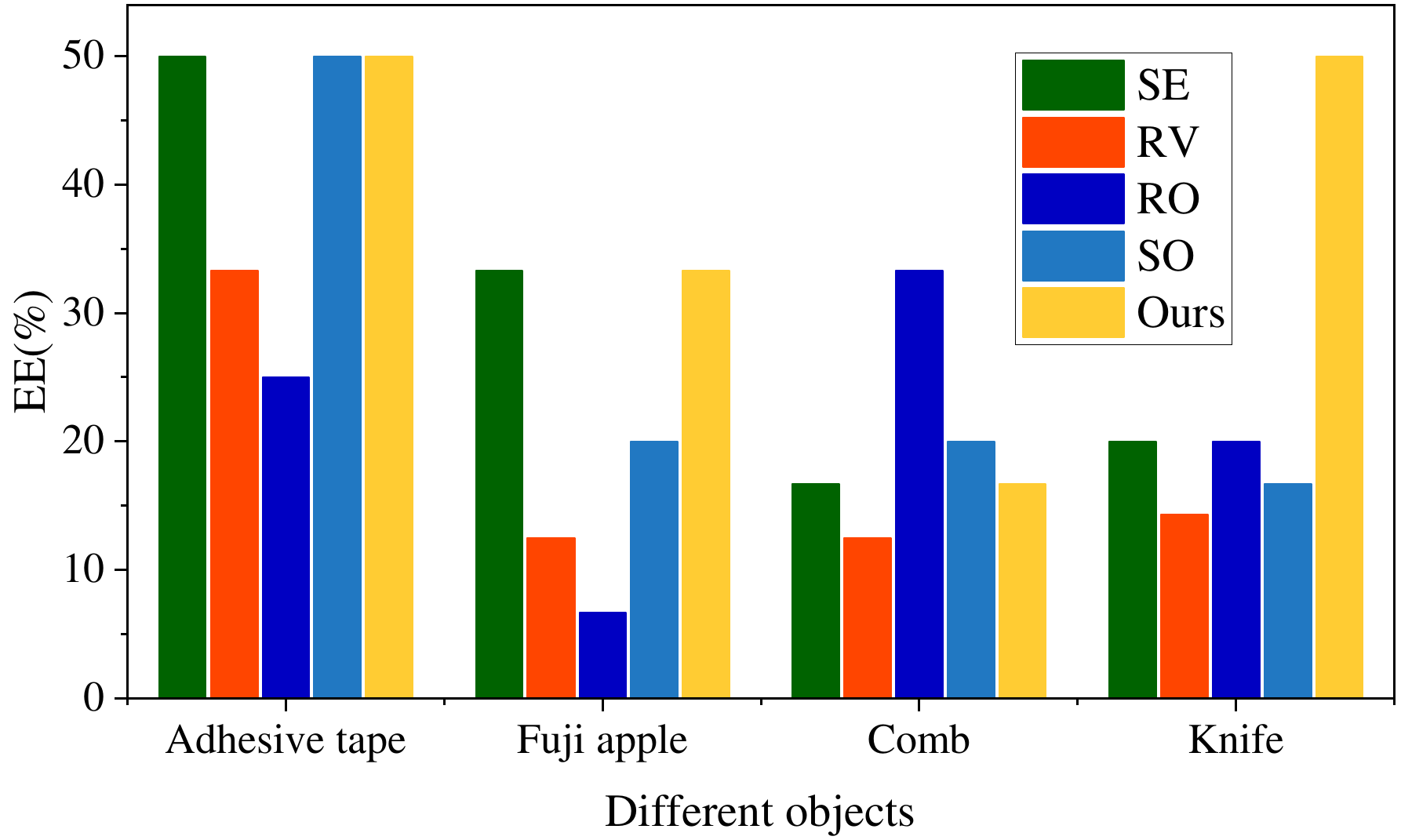}
    \caption{Different sampling approaches for the cross-domain setting.}
    \label{fig:res01}
\end{figure}
\begin{table}
    \centering
    \caption{The results of different adaptation techniques on cross-domain datasets. GT denotes that we utilize the annotated dataset to finetune the network.}
    \begin{tabular}{c|ccc}
\toprule
    Settings&Method& Accuracy (\%)&$\Delta\uparrow$ \\
    \cmidrule{1-4}
         \multirow{1}{*}{Baseline}&GGCNN&18.52\%&-\\
    \cmidrule{1-4}
         \multirow{1}{*}{Finetune}&GT&70.37\% &51.85\%\\  
    \cmidrule{1-4}
         \multirow{2}{*}{ETA}&Ours+Single-view&33.33\%&14.81\%\\
         &Ours+Multi-view&59.26\%&40.74\%\\    
    \bottomrule
    \end{tabular}
    \label{tab:main01}
\end{table}

To assess the effectiveness of our proposed framework (ETA), we conducted a comparative analysis involving a strong baseline GGCNN and a finetuned grasp detection network. The evaluation encompasses both cross-domain and same-domain scenarios. In the cross-domain setting, the networks are pre-trained using the Cornell dataset and subsequently applied in a distinct domain. Conversely, the same-domain setting involves networks pre-trained with an extracted real-world dataset and deployed in the same environment. Besides, we also compare our method with two different variants, \ie single-view for only utilizing the first viewpoint image content and multi-view for actively exploring the sufficient image content. The results are reported in Table~\ref{tab:main01} and Table~\ref{tab:main02}. 
\begin{figure}
    \centering
    \includegraphics[width=.85\linewidth]{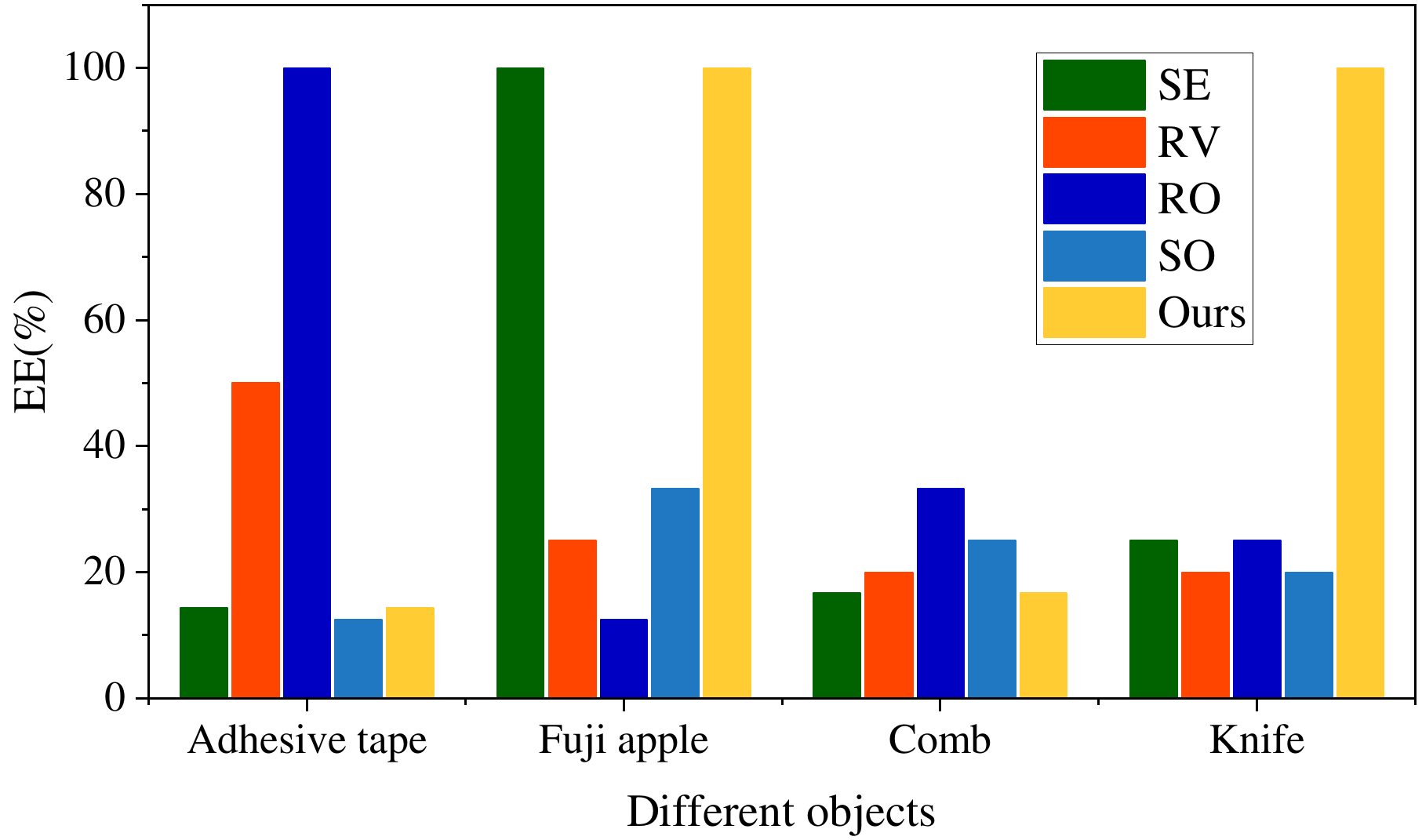}
    \caption{Different sampling approaches for the same-domain setting.}
    \label{fig:res02}
\end{figure}
\begin{table}
    \centering
    \caption{The results of different adaptation techniques on same-domain datasets.}
    \begin{tabular}{c|ccc}
\toprule
    Settings&Method& Accuracy (\%)&$\Delta\uparrow$ \\
    \cmidrule{1-4}
         \multirow{1}{*}{Baseline}&GGCNN&37.04\%&-\\
    \cmidrule{1-4}
         \multirow{1}{*}{Finetune}&GT&77.78\%&40.74\% \\   
    \cmidrule{1-4}
         \multirow{2}{*}{ETA}&Ours+Single-view&55.56\%&18.52\%\\
         &Ours+Multi-view&62.90\%&25.86\%\\
    \bottomrule
    \end{tabular}
    \label{tab:main02}
\end{table}

We can observe that: 1) our method obtains superior performance and outperforms GGCNN by a large margin (59.26\% \textit{v.s.} 18.52\% for the cross-domain setting, 62.90\% \textit{v.s.} 37.04\% for the same-domain setting). 2) although GGCNN achieved high accuracy in Cornell as reported in~\cite{morrison2018closing}, there is still a significant performance decline in practical applications. We speculate that this may be due to a significant domain gap in environmental information between real scenes and datasets. 3) As for the comparison with a finetuned method, we find that there is still a significant performance gap (70.37\% $v.s.$ 59.26\%, 77.78\% $v.s.$ 62.90\%). However, the finetuned method is trained with the labeled samples. The process is a time-consuming and laborious task, making it difficult for robots to achieve long-term autonomous learning.
\subsection{Ablation Studies}
\paragraph{Results of Different Sampling Approaches}
To reveal the effectiveness of the knowledge retrieval module for the initial viewpoint setting, we compare our methods with three different strategies, \ie Sequential exploration (SE), Random exploration of viewpoints (RV), Random exploration of observations (RO), Sequential exploration of observations and random exploration of viewpoints (SO). Four candidate objects, adhesive tape, Fuji apple, comb, and knife, are selected to verify the effectiveness of our knowledge retrieval module. The comparison results across cross-domain and same-domain settings are illustrated in Fig.~\ref{fig:res01} and Fig.~\ref{fig:res02}, respectively.

\begin{figure*}
    \centering
    \includegraphics[width=.8\linewidth]{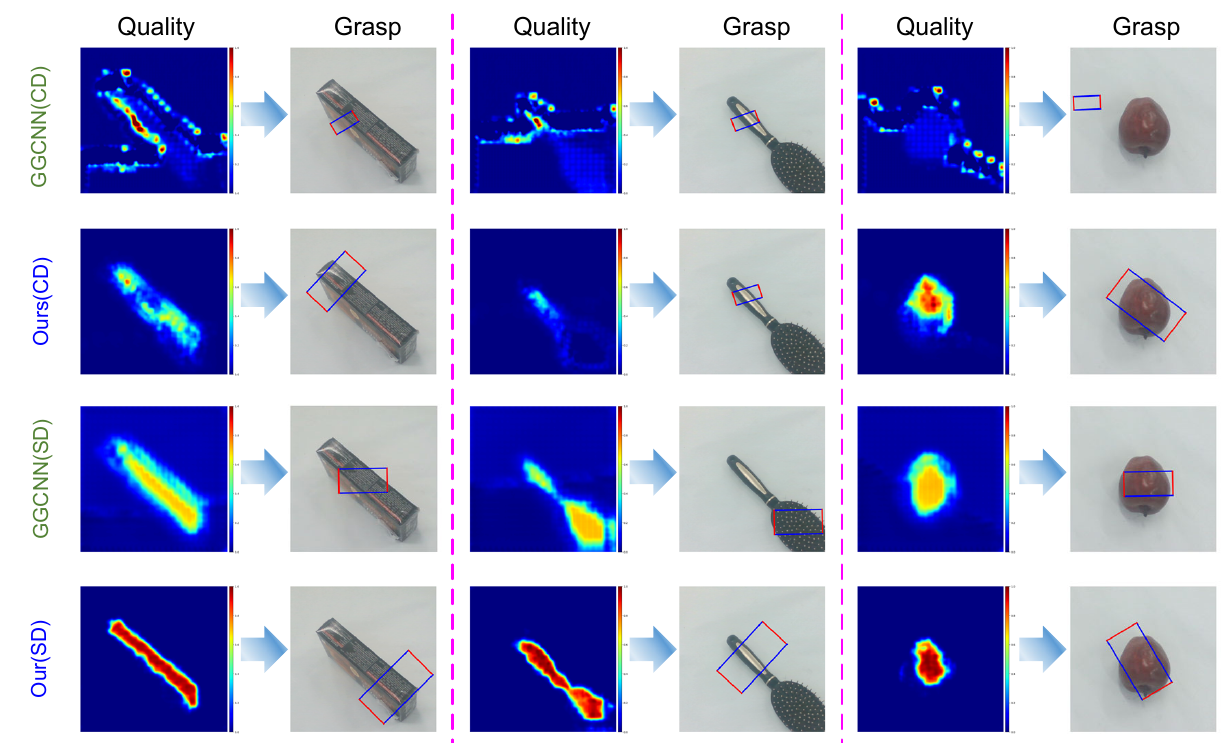}
    \caption{Qualitative comparison results between GGCNN and embodied test-time adaptation. CD denotes the cross-domain setting, and SD represents the same-domain setting.}
    \label{fig:enter2}
\end{figure*}
\begin{figure}
    \centering
    \includegraphics[width=0.85\linewidth]{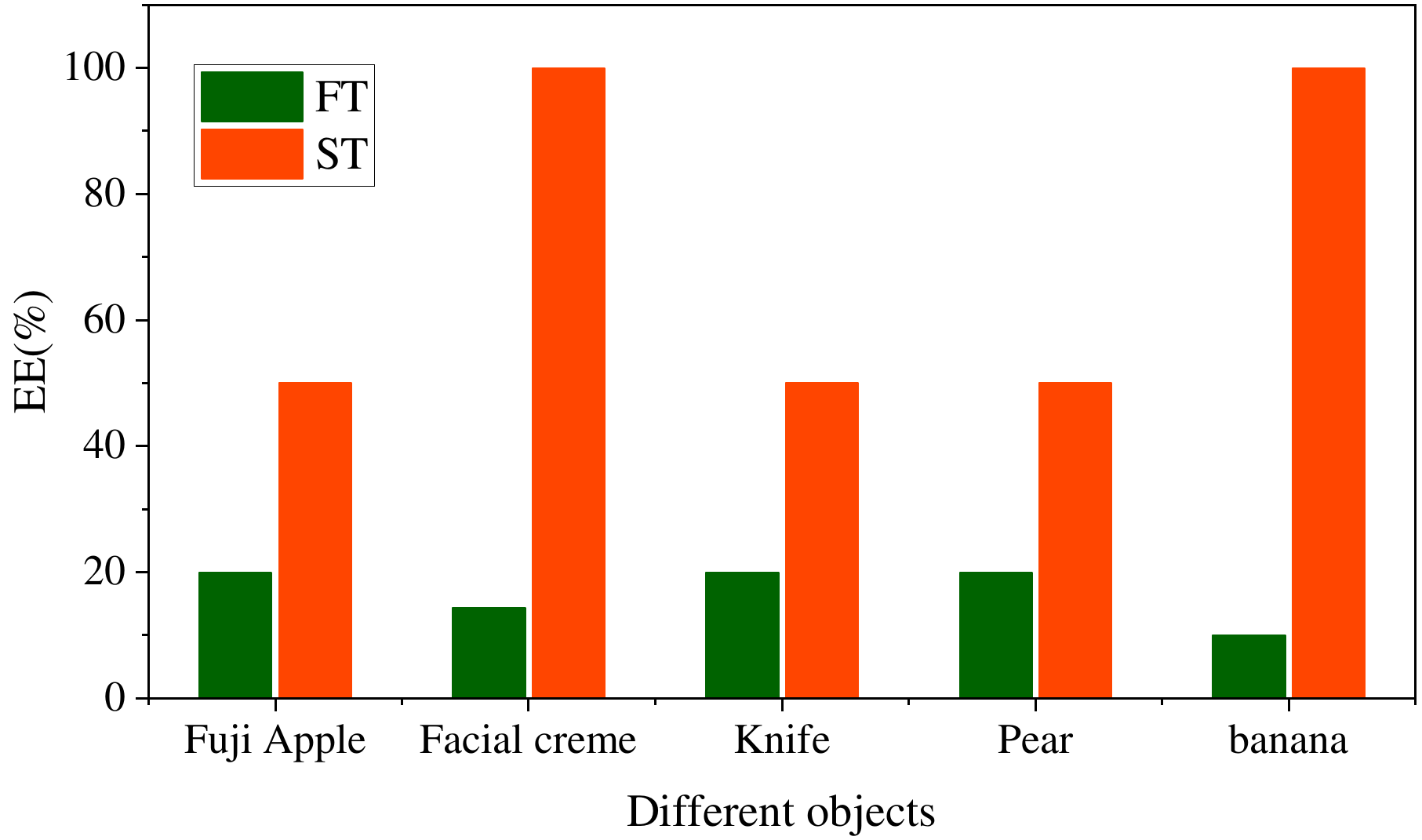}
    \caption{Results of two grasp processes before and after test-time adaptation for the cross-domain setting. \hl{FT and ST denote seeing the object for the first time and second time}, respectively (the same as below).}
    \label{fig:firstandsecond}
\end{figure}
As can be observed from the results, we can find that, during the execution of a grasping operation on a novel object without relevant knowledge references (\eg adhesive tape, Fuji apple, and comb), our method employs a sequential retrieval strategy, which does not consistently yield optimal performance. Conversely, although certain random perspectives facilitate the identification of grasping positions, they lack robustness. However, when the object possesses applicable knowledge (\eg, comb and knife), the model conducts a search to establish the initial observation. At this juncture, a notable enhancement in the model's performance is evident. For example, both the comb and the knife are categorized as handle objects. After learning how to grasp the comb, the model can easily determine the potential optimal observation of the knife. This further confirms the importance of knowledge retrieval for active exploration.


%

\paragraph{Results of Different Threshold $\epsilon$}
The threshold $\epsilon$ affects whether the robot performs grasping operations. To verify the effect of our method performance on different $\epsilon$, we conduct experiments in both cross-domain and same-domain settings, where $\epsilon$ is set to 3, 4, and 5. The results are reported in Table~\ref{tab:threshold}, showing that the method achieves the best performance when the value is set to 4. We speculate that stricter conditions (e.g., $\epsilon$=3) may yield fewer effective samples, while looser conditions (e.g., $\epsilon$=5) could introduce noise. Consequently, we empirically set the threshold $\epsilon$ to 4 for the experiments in this paper.

\paragraph{Qualitative Results}
\begin{figure}
    \centering
    \includegraphics[width=0.85\linewidth]{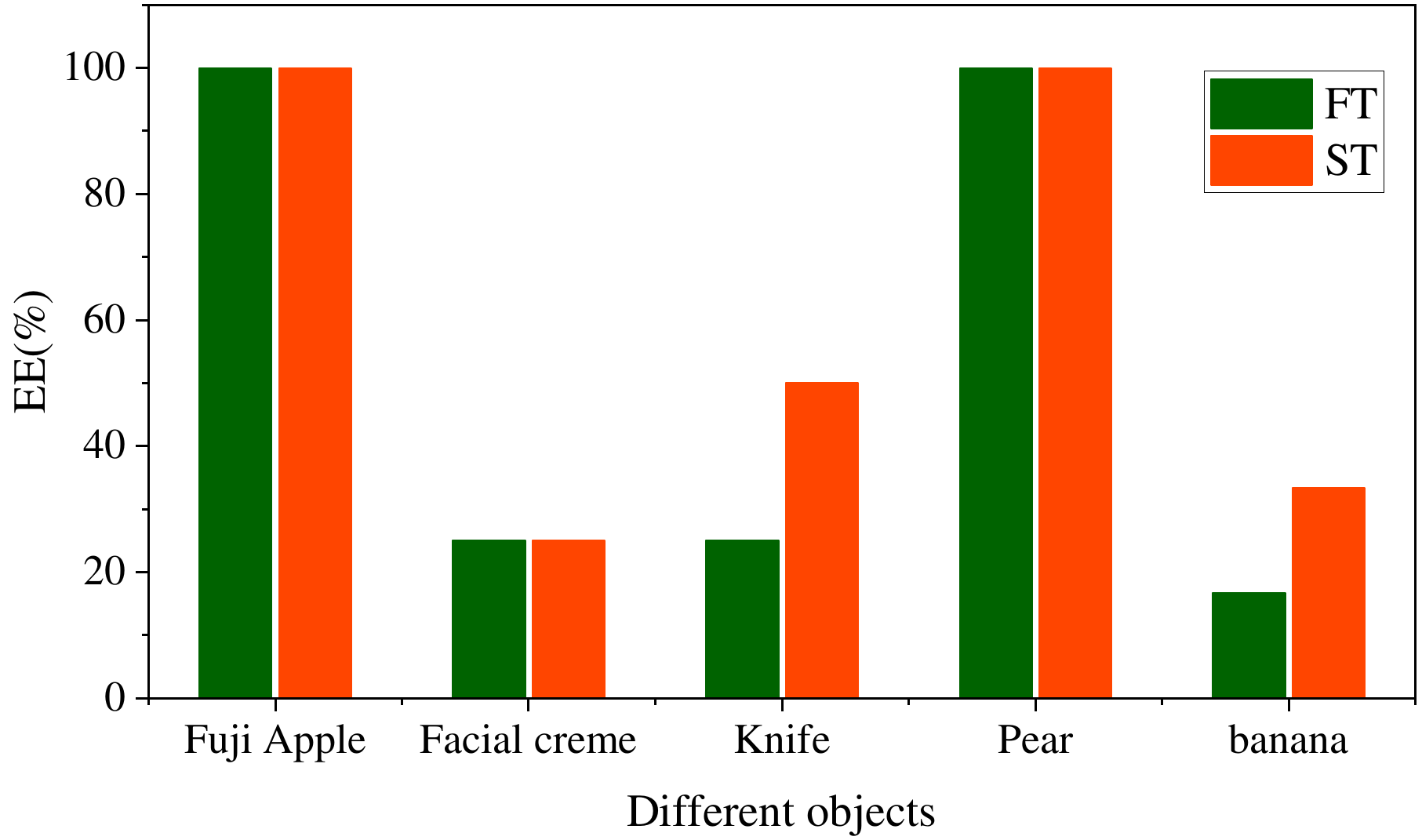}
    \caption{Results of two grasping before and after test-time adaptation for the same-domain setting.}
    \label{fig:firstandsecond2}
\end{figure}


\begin{table}
    \centering
    \caption{Result comparison with different threshold values for $\epsilon$.}
    \begin{tabular}{cccc}
    \toprule
         Threshold&Domains&Accuracy(\%)&Average  \\
         \cmidrule{1-4}
        \multirow{2}{*}{ $\epsilon$=3.0}& Cross-Domain&59.26\%&\multirow{2}{*}{59.26\%}\\
        &Same-Domain&59.26\% \\
         \cmidrule{1-4}
        \multirow{2}{*}{ $\epsilon$=4.0}& Cross-Domain&59.26\%&\multirow{2}{*}{61.08\%}\\
        &Same-Domain&62.90\% \\
         \cmidrule{1-4}
        \multirow{2}{*}{ $\epsilon$=5.0}& Cross-Domain&48.15\%&\multirow{2}{*}{51.86\%}\\
        &Same-Domain&55.56\% \\
        \bottomrule
    \end{tabular}
    \label{tab:threshold}
\end{table}

Our embodied test-time adaptation aims to improve the generalization performance of grasping skills for robots in unseen environments based on their own capabilities with newly collected samples. To demonstrate the effectiveness of our method, we visualize the qualitative results in Fig.~\ref{fig:enter2}. We can observe that our framework can effectively improve the model's grasping detection performance for cross-domain and same-domain settings. Even when the model predicts inappropriate results, as shown in the third column of GGCNN(CD), our method can still predict the right grasping rectangle. These results further exemplify the effectiveness of our approach in addressing the domain gap.
\paragraph{Results of Same Object Grasping}
To validate the effectiveness of the proposed test-time adaptation, we conduct experiments on the same objects both before and after test-time adaptation across cross-domain and same-domain settings. The results are illustrated in Fig.~\ref{fig:firstandsecond} and Fig.~\ref{fig:firstandsecond2}. \hl{Notably, FT also denotes the exploration process of a new object}. As can be observed from the figures, our approach demonstrates a notable enhancement in grasping accuracy upon encountering an object for the second time. Moreover, the improvements are significant within the same-domain setting compared with the cross-domain setting.

\section{Discussions}
\hl{Although our framework demonstrated superiority on real-world datasets across same-domain and cross-domain settings, our method struggles to deal with unfamiliar objects. This may be due to the fact that our method possesses no relevant knowledge of previous objects, which is also the reason why sometimes the model's performance is not optimal. In the future, we plan to utilize LLMs to generate relevant grasping knowledge to further improve exploration efficiency and apply it to various kinds of robots~\cite{zhouziyang,bjelonic2023learning}.}
\section{Conclusion}
In this paper, we propose an embodied perception framework with knowledge infusion for robotic grasping detection during the test-time adaptation. The framework aims to improve the generalization performance of grasping skills for robots in new environments based on their own active exploration capabilities. Importantly, to evaluate the quality of the grasping detection results, we introduce an assessment criteria based on the robot's embodied parameter to eliminate human intervention and realize self-supervised test-time adaptation. Besides, we construct a knowledge pool for the robot to acquire an initial optimal viewpoint, thus improving the efficiency of the exploration. Finally, we demonstrate our framework on real-world datasets across same-domain and cross-domain settings. 

\bibliographystyle{IEEEtran}
\bibliography{refs}

\vfill

\end{document}